\documentclass[letterpaper, 10 pt, conference]{ieeeconf}  

\usepackage{comment}
\usepackage{amsmath,amssymb}
\usepackage{color}
\usepackage{url}
\usepackage{hyperref}
\usepackage{graphicx}
\usepackage{amsmath,amssymb}
\usepackage{tikz}
\usepackage{cite}
\usepackage{floatrow}
\usepackage{comment}
\usepackage{amsmath,amssymb} 
\usepackage{color}
\usepackage{floatrow}
\usepackage{array}
\usepackage{float}
\usepackage{mathtools}
\usepackage{booktabs}
\floatstyle{plaintop}
\usepackage[export]{adjustbox}

\restylefloat{table}
\newcolumntype{M}{>{\centering\arraybackslash}m{\dimexpr.22\linewidth-2\tabcolsep}}
\newcolumntype{S}{>{\centering\arraybackslash}m{\dimexpr.05\linewidth-5\tabcolsep}}
\DeclareMathOperator{\sign}{sgn}

\newenvironment{tightitemize} 
{\vspace{-\topsep}\begin{itemize}\itemsep1pt \parskip0pt \parsep0pt}
{\end{itemize}\vspace{-\topsep}}

\DeclareMathOperator*{\argmin}{arg\,min}

\IEEEoverridecommandlockouts                              

\title{\LARGE \bf
Memory Maps for Video Object Detection and Tracking on UAVs 
}

\author{Benjamin Kiefer$^{1}$, Yitong Quan$^{1}$ and Andreas Zell$^{1}$
\thanks{$^{1}$All authors are with the Faculty of Computer Science,
        University of Tuebingen, Germany.
        {\tt\small prename.surname@uni-tuebingen.de}}%
}

\begin{document}

\maketitle
\thispagestyle{empty}
\pagestyle{empty}

\begin{abstract}
   This paper introduces a novel approach to video object detection detection and tracking on Unmanned Aerial Vehicles (UAVs). By incorporating metadata, the proposed approach creates a memory map of object locations in actual world coordinates, providing a more robust and interpretable representation of object locations in both, image space and the real world. We use this representation to boost confidences, resulting in improved performance for several temporal computer vision tasks, such as video object detection, short and long-term single and multi-object tracking, and video anomaly detection. These findings confirm the benefits of metadata in enhancing the capabilities of UAVs in the field of temporal computer vision and pave the way for further advancements in this area.
\end{abstract}

\section{Introduction}


When we make predictions about the presence and location of objects, we have an internal understanding of our surrounding world: implicitly, we know where we are in relation to the object and we know about the topology of a given scene. This internal understanding of the surrounding geometry allows us to reason robustly about the existence and location of objects. 
Furthermore, while we make detection errors when shown ambiguous static scenes, over time, we are able to strengthen our belief about our predictions. This is due to slight changes in appearance caused by different view points, slight movement of objects or just by integrating our predictions over a certain period of time.

We argue that this awareness of our surrounding is particularly important in aerial scenarios, where we need to reason about our environment in the presence of many uncertainties, caused by the smallness of objects. Motivated and inspired by this human-based analogy, in this work, we aim to improve several computer vision tasks for object detection and tracking on UAVs.

\begin{figure}
    \centering
    \includegraphics[width=1\textwidth]{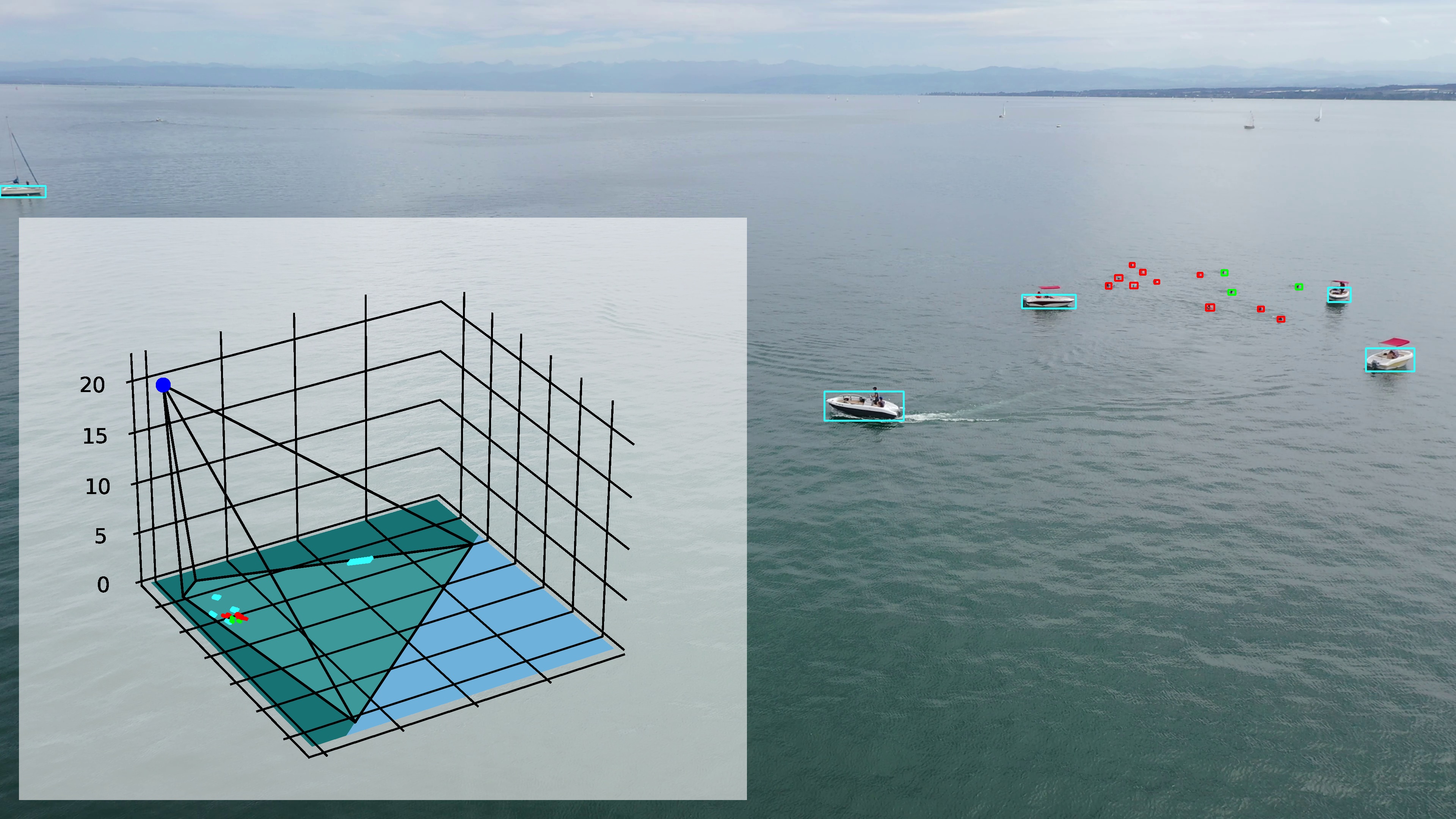}
    \caption{Illustration of predictions and the corresponding 3D geometry/topology. We  aggregate information over time, taking the correct 3D geometry into account.}
    \label{fig:intro_img}
\end{figure}

Conventional techniques for object detection and tracking from UAV perspectives ignore the intrinsic geometry and topology present in UAV-generated imagery, frequently relying on off-the-shelf methods designed for COCO-like scenarios or with slight modifications, see e.g. \cite{mittal2020deep}. This makes it difficult for these methods to aggregate uncertain predictions over time, since movements in image space cannot correctly be tracked and, hence, features are hard to be associated temporally.

We argue that this shortcoming can easily be mitigated by leveraging freely available sensors onboard the UAV. GPS alongside compass and IMU measurements allow us to reason about our and the objects' locations in 3D coordinates. In turn, this allows us to create a \emph{temporal memory map} of previous predictions, so that we can aggregate information over time in a geometrically sensible way, resulting in more robust predictions.

In this work, we show how correctly considering the 3D geometry allows us to propose a temporal memory map that results in more robust predictions. In particular, our contributions are as follows:
\vspace{3mm}

\begin{tightitemize}
    
    \item We derive mathematical formulas detailing the 3D geometry around a UAV.
    \item Leverage metadata, we propose a memory map to robustify several temporal computer vision tasks, such as video object detection, tracking and anomaly detection.
    \item We capture PeopleOnGrass-Video, a benchmark featuring 4K 30 FPS manually annotated video, featuring precise metadata labels and make it publicly available. Furthermore, we capture and annotate more data for multiple further experiments and release it publicly.
    \item We show in multiple experiments on diverse benchmarks the utility of our method.
    
\end{tightitemize}


\section{Related Work}

\begin{figure*}
    \centering
        
        \includegraphics[trim=0 0 0 0,clip,width=0.49\textwidth]{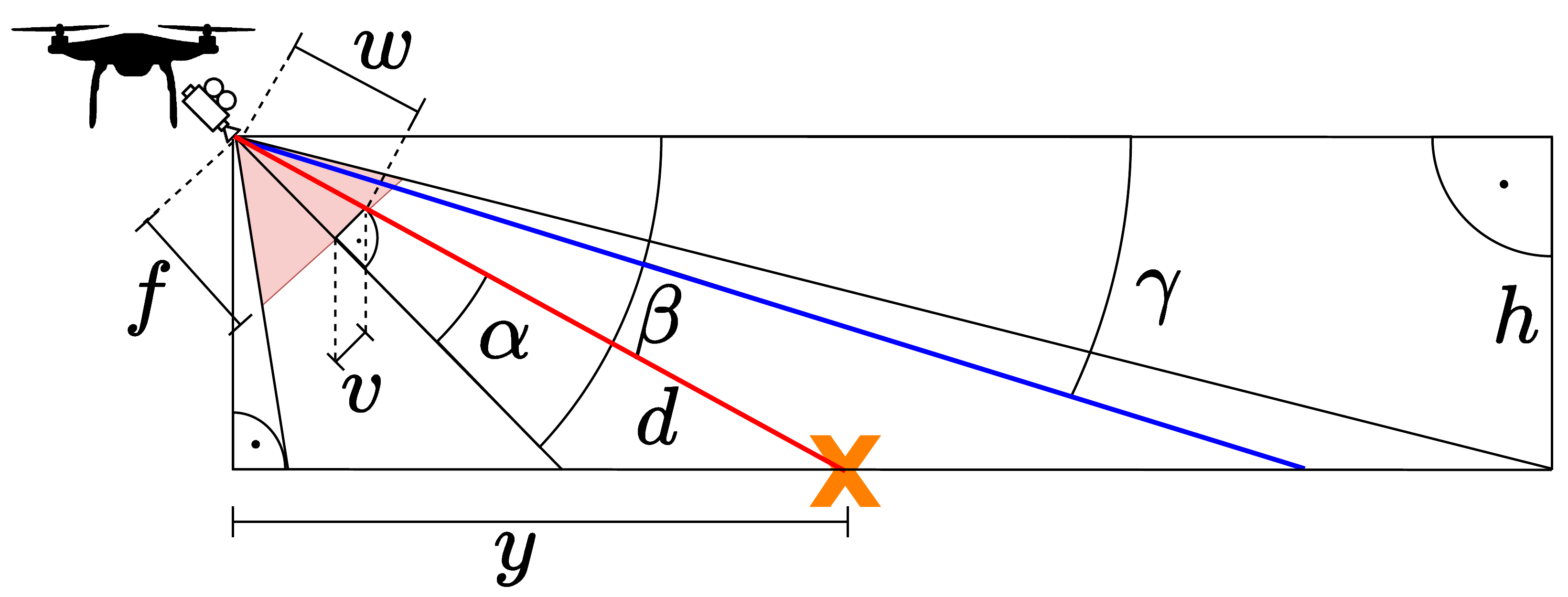}
                \includegraphics[trim=0 0 0 0,clip,width=0.49\textwidth]{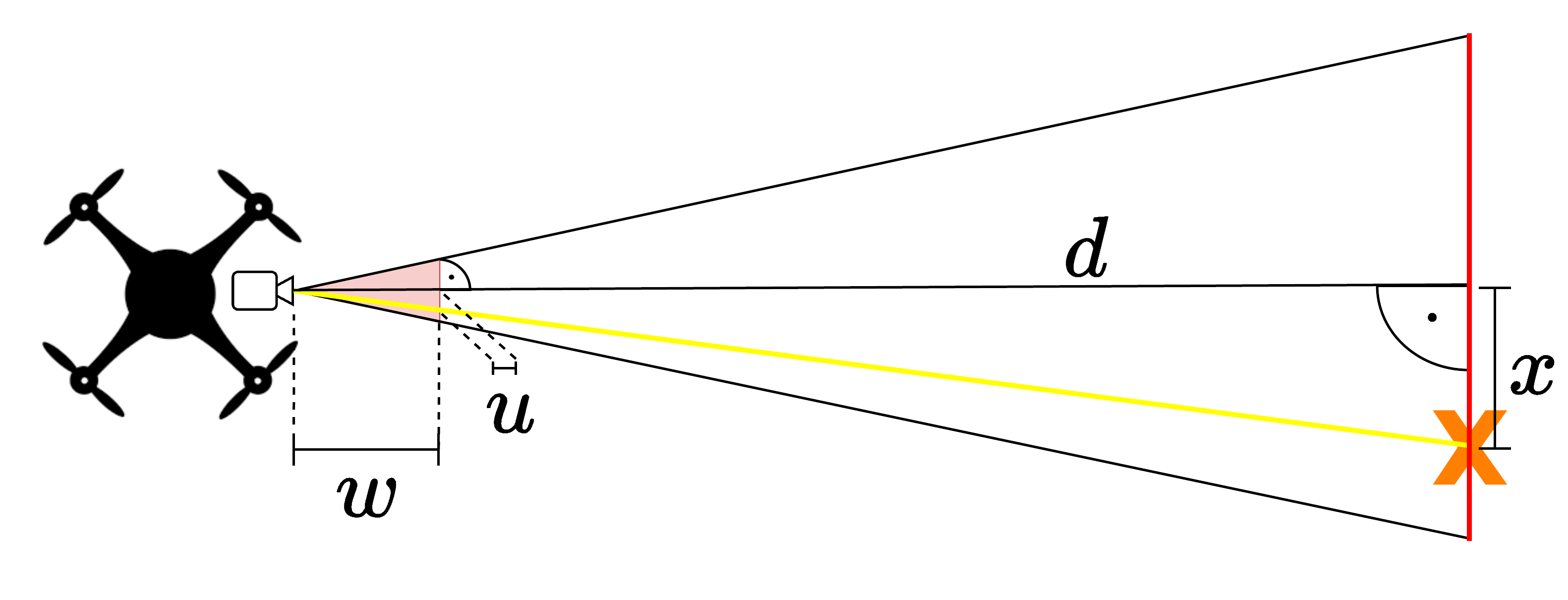}
    \caption{Illustration of variables and formulas to estimate the vertical distance $y$ (red lines) and the horizontal distance $x$ (yellow lines, illustrated on bottom left) from the UAV to the swimmer (orange cross).}
    \label{fig:mathstuff}
\end{figure*}

Although video object detection (VOD) on UAVs has become more relevant, with many applications emerging and many network architectures becoming fast enough to deploy on embedded devices, there is still no consensus on which VOD approach seems to be the most promising. Broadly speaking, there are optical flow-based networks, memory networks and tracking-based networks \cite{wu2021deep}. While memory networks, such as STDnet-ST \cite{bosquet2021stdnet}, achieve high accuracies, their heavy architectures prohibit their deployment on embedded devices. Optical flow-based networks are faster, but their benefit over single-frame object detectors is limited \cite{corsel2023exploiting}. Most tracking-based networks first do single-frame object detection and use the association for the detection reciprocally \cite{luo2019detect}. In fact, the best two VOD models of the last VisDrone-VID challenge were single-frame methods, entirely ignoring the temporal domain \cite{zhu2019visdrone}.

Furthermore, all common video object detectors on UAVs ignore the underlying 3D geometry of the scene and only operate in image space \cite{wu2021deep}. While there is much research focusing on geolocation, it is only focused on obtaining world coordinates of objects for downstream tasks, such as following a target \cite{zhao2019detection}. Similarly, occupancy networks and other mapping approaches aim at obtaining a map for mapping or scene understanding \cite{wei2022three}.

As opposed to these works, we aim to leverage metadata to build a memory map in GPS space that \emph{improves the video object detection performance}. Since we only rely on freely available metadata onboard the UAV, this approach is viable for small UAVs with a standard RGB camera in low-cost settings (compare to active geolocation via laser \cite{yang2019high}).








\section{Deriving Formulas for 3D Geometry}
\label{sec:mathsection}




\begin{figure}
    \centering    

        \includegraphics[trim=0 0 0 0,clip,width=1\textwidth]{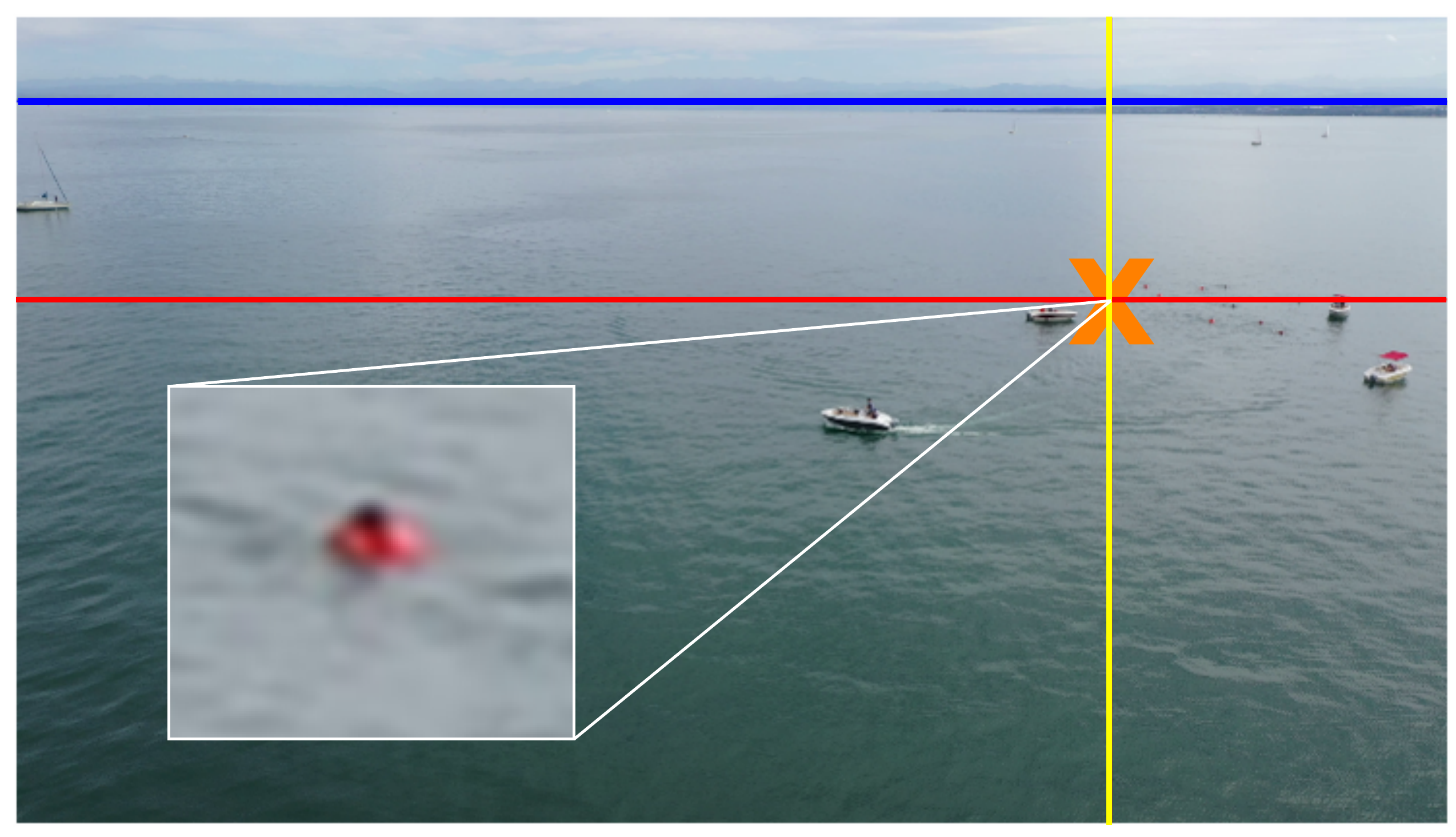}
    \caption{Corresponding image example for the computation in Figure \ref{fig:mathstuff}. Here, we have that $h=20m$ and $\beta=15^\circ$ resulting in $y=110m$ and $x=25m$. This means, that the swimmer is $110m$ in front and $25m$ to the right of the UAV as measured on the ground. Note that we also estimate the horizon line as indicated by the blue line.}
    \label{fig:fig:how_far_away}
\end{figure}

Figure \ref{fig:mathstuff} illustrates that we would like to know the GPS position of the swimmer as indicated by the orange cross. How do we obtain that? First, we discuss how to obtain relative coordinates to the UAV, then how to use these to obtain actual world coordinates via passive geolocation.

\subsection{Relative Coordinates}

We consider a mathematical perspective projection camera model since this resembles the common use-case for cameras on UAVs. We consider our camera to have a focal length $f$ given in pixels. For simplicity, we assume our image to be of 4K (3840x2160) resolution. Other resolutions follow an analogous derivation. Our camera is looking down at an angle of $\beta$, which is the variable gimbal angle. The gimbal also balances a potential UAV roll angle, so that we assume there to be a zero camera roll angle.

First, we estimate the on-ground distance from the UAV to the red line, which we denote as $y$ (see Figure \ref{fig:mathstuff}). We assume the ground to be flat, i.e. we ignore elevation change and the curvature of the earth. The latter is a reasonable assumption for heights that are typical for UAV missions \cite{bohren1986altitude}, while the first may make a difference in very dynamic terrains.

Using the focal length $f$ and the y-axis pixel position offset $v$ from the horizontal center pixel line of the image, we compute $y$ by computing the angle $\alpha$ via

\begin{equation}
    \alpha = \arctan \left(\frac{v}{f}\right).
\end{equation}

This allows us to compute $y$ via 

\begin{equation}
    y = \tan (90^\circ - (\beta - \alpha))h.
\end{equation}

Having the vertical ground-distance to the object of interest, we compute the horizontal ground-distance $x$ by looking from the top as indicated on the bottom left of Figure \ref{fig:mathstuff}. For that, we need the variables $d$ and $w$, which both are easily computed via Pythagoras' theorem, which then yield $x$:

\begin{align}
        d = \sqrt{h^2+y^2}, \text{    }        w = \sqrt{f^2+v^2},\text{    } x = \frac{u}{w}d.
\end{align}


For simplicity, we ignored the sign of $x$, which indicated whether we are on the left or on the right of the vertical center line. Naturally, in the actual implementation, we incorporate that. Likewise for the special cases when an object of interest is behind the UAV, which happens for gimbal angles close to $90^\circ$.

Furthermore, we compute the horizon line in image space following a similar derivation as \cite{kiefer2023fast}. Ignoring the effect of atmospheric refraction, we estimate the distance to the horizon line $l$ as a function of the height of the UAV as 

\begin{equation}
    l=3.57h^{1/2}. 
\end{equation}

This approximation is fairly accurate for heights that are typical for our scenarios (far below 1000m) \cite{bohren1986altitude}. We compute the angle $\gamma$ to the horizon via 

\begin{equation}
    \gamma= \arcsin (h/l).
\end{equation}

Using the focal length $f$ and the camera gimbal pitch $\beta$, we compute the camera perspective projection of the horizon on the image plane, which yields the height offset $o$ in pixels to the horizontal center line of the image plane as 

\begin{equation}
    o=\tan (|\gamma-\beta|) \cdot f \cdot \sign (\gamma-\beta).
\end{equation}

Finally, we need to truncate $o$ to be within the range of the number of allowable horizontal pixels. We refer to \cite{kiefer2023fast} for a discussion on accuracy. Lastly, note if there is a lens distortion, then we undistort the image before running the above algorithm.

\subsection{Absolute Coordinates}

Using the relative coordinates of an object ($x$- and $y$- ground-distances to UAV), we compute its GPS coordinates based on the UAV's GPS coordinates as follows. Given the camera heading angle $\theta$, we compute the rotation matrix and rotate the relative coordinates of an object to obtain

\begin{equation}
    \begin{bmatrix}
        x_r \\ y_r \\ 1
    \end{bmatrix} = \begin{bmatrix}
\cos (\theta) & \sin (\theta) & 0\\
\sin (\theta) & \cos(\theta) & 0\\
0 & 0 & 1
\end{bmatrix}    \begin{bmatrix}
        x \\ y \\ 1
    \end{bmatrix}.
    \label{equation:coordinate_transfer}
\end{equation}

Finally, we map the relative coordinates to GPS coordinates via
\begin{align}
    la^{object} &= la + \frac{y_r}{r} \frac{180}{\pi}, \\
    lo^{object} &= lo + \frac{x_r}{r} \frac{180}{\pi} \frac{1}{\cos (lat \ \pi / 180)}.
\end{align}

We note that we compute the earth radius based on the latitude to account for the slight ellipsoid nature of the earth. Lastly, note that we can reverse the computation to map from 3D to 2D, i.e. the image space (see code that we'll make publicly available).

\section{Method: Temporal Memory}







Classical grid maps in robotics, such as occupancy grid maps, divide the
environment into single grid cells and estimate the occupancy
probability for each cell \cite{nuss2018random}. We also aim to build temporary memory maps but in the context of modeling dynamic objects of interest. While there are several works extending occupancy grid maps to account for dynamic objects, e.g. \cite{chung2010slammot}, our focus is on leveraging maps to improve the downstream performance of computer vision detection and tracking algorithms over time. Hence, we propose a simple class of what we denote \emph{temporal memory maps}.

\subsection{Map Representation}

We consider a UAV-centric map with a fixed size context window, i.e., given the latitude $la_t$ and longitude $lo_t$ of the UAV at discrete time step $t$, the memory map is defined to be
\begin{align}
 M_t=
        \left\{ \begin{pmatrix}
            x\\y
        \end{pmatrix} \bigg| \begin{matrix}
x\in \{ la_t-\frac{c}{2}+\frac{j}{n}  | j \in 0,...,n (la_t+\frac{c}{2})\} \\
y\in \{ lo_t-\frac{c}{2}+\frac{j}{n} | j \in 0,...,n (lo_t+\frac{c}{2})\}
\end{matrix} \right\}     ,
\end{align}
which denotes a quadratic, north-oriented map around $la_t$ and $lo_t$ of edge size $c$ with $n^2$ equidistantly spaced elements. Note that $c$ and $n$ should be problem-dependent reasonable quantities. This treatment makes sure that the surrounding environment of the UAV can be considered at all times. Note that $c$ should be large enough to account for the relevant field of view of the UAV and $n$ should be large enough to have sufficient resolution. Furthermore, the map will forget information that is outside of the context window, which needs to be considered in downstream applications. 

From here, the exact procedure of how to update and leverage the map differs between the computer vision tasks. Broadly speaking, all methods define a memory map on $M_t$, indicating likelihoods of object locations.

\subsection{Video Object Detection}

\begin{figure}
    \centering
    \includegraphics[width=1\textwidth]{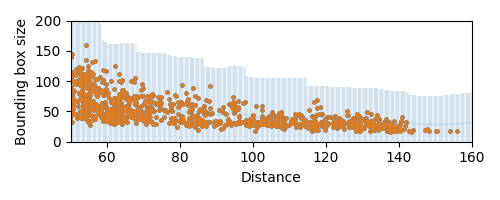}
    \caption{Distance (x-axis) to size (y-axis) relation for all swimmers in SeaDronesSee object detection train set (orange dots). The blue area denotes the accepted area of sizes.}
    \label{fig:distancesize}
\end{figure}

After analyzing single-image object detectors, we found that most trained object detectors can detect almost all objects before the non-maximum suppression (NMS) stage, which in our case is doing both, removing duplicate boxes and low-confidence ones. Perhaps unsurprisingly, this is due to object detectors distributing many anchors across the image, from which potential objects are regressed. However, detections tend to cluster around actual objects, thereby achieving theoretical recall values of close to one.

\begin{figure}
    \centering
    \includegraphics[width=1\textwidth]{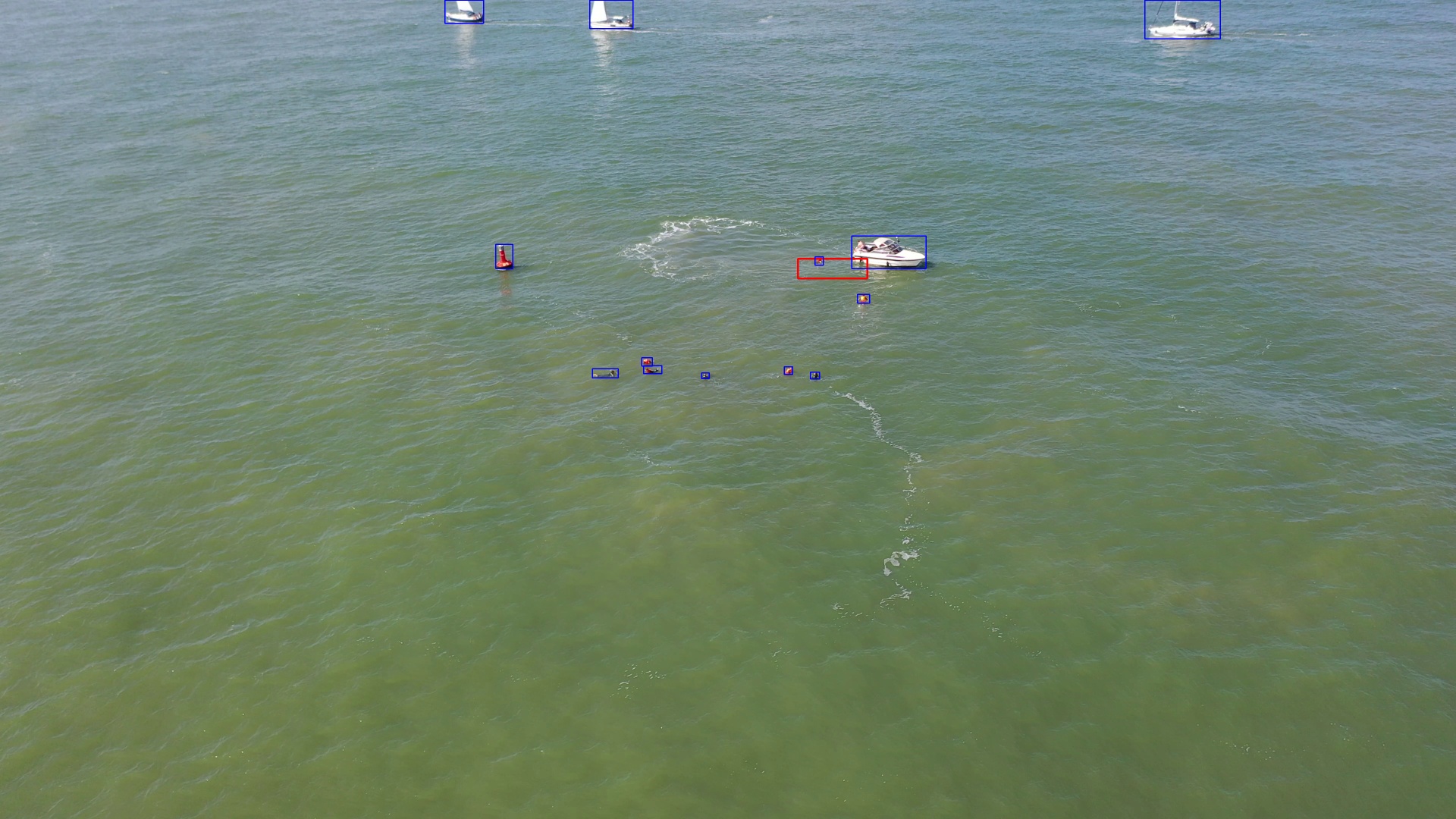}
    \caption{Blue boxes are true positive detections, the red box is a false positive swimmer detection that was successfully filtered out by analyzing the distance-size relation.}
    \label{fig:distance_size_relation}
\end{figure}

An immediate observation yields that we can filter out predictions based on their sizes. Common object detectors employ multiple stages, responsible for multiple scales. Often, the detectors output predictions, whose sizes are far off from any reasonable quantities. By looking at the distance $d$ and the class indicator of any predicted object, we establish a distance-to-size (size measured by diameter of the box) relation and analyze it on the SeaDronesSee object detection train set (see Figure \ref{fig:distance_size_relation}). For each class separately, we perform a Gaussian process regression on these data points. This yields a mean function $m(x)$ and a covariance function $cov(x)$. We scale the covariance function by a factor dependent on the distance from the maximal to the minimal bounding box size in a small interval around $x$. Finally, we check whether a new prediction is inside the resulting accepted area, and if not, we discard that prediction (see Figure \ref{fig:distancesize}). Applying this procedure on the three best models of the SeaDronesSee Object Detection v2 challenge, Table \ref{table:removedboxes} shows that we can already remove a small amount of FPs.

\begin{table}	
	\begin{center}		
		\begin{tabular}{lrrrr}
                \toprule
			Model name & \#B & Rem.FP & Rem.TP & \% \\
			\midrule
                Maritime-VSA \cite{kiefer20231st} & 674K & 8,087 & 24 & 1.2 \\
                DetectoRS \cite{kiefer20231st} & 838K & 6,705 & 67 & 0.8\\
                YOLOv7-Sea \cite{kiefer20231st} & 378K & 4,926 & 54 & 1.3 \\
                \bottomrule
		\end{tabular}
	\end{center}
	\caption{Numbers of removed (Rem.) boxes (B) for best three models trained and tested on SeaDronesSee Object Detection v2.}
    \vspace{-3mm}
	\label{table:removedboxes}
\end{table}

\begin{figure}
    \centering
    \includegraphics[width=1\textwidth]{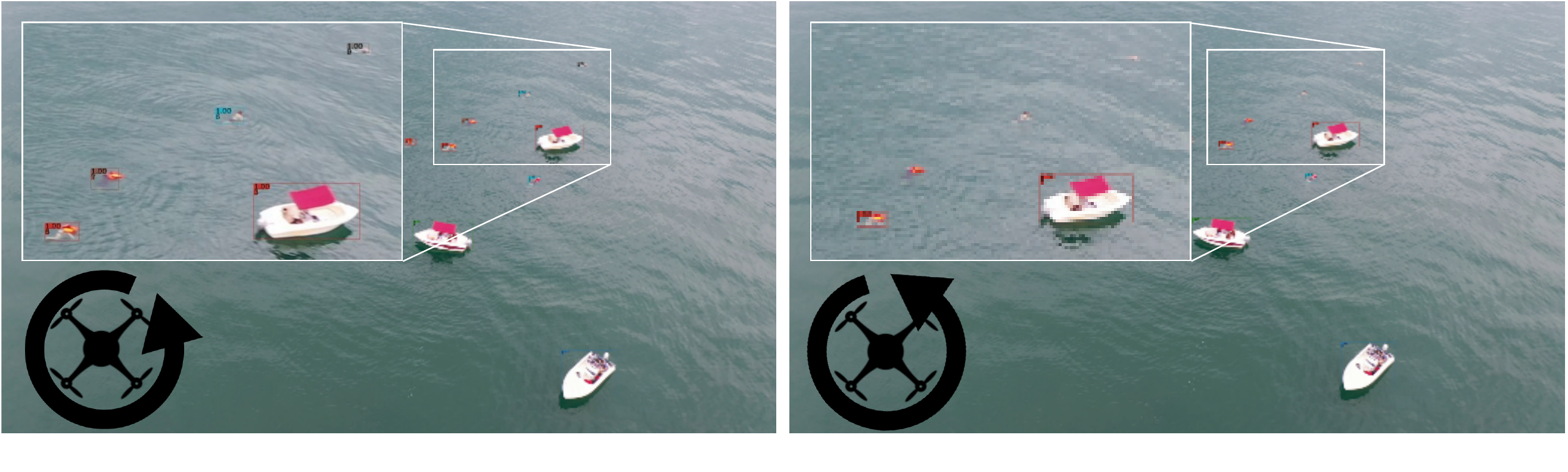}
    \caption{A heading change of the camera results in three swimmers (detected in left frame) being lost in subsequent frames (right frame) (DeepSORT \cite{kiefer20231st}).}
    \label{fig:heading_rotations}
\end{figure}

Nevertheless, there needs to be a suppression stage because the high recall still comes at the cost of many false positive bounding boxes. We argue that robust detection can be achieved by temporally and geometrically sensible aggregation of many of the low-confidence detections that cluster around actual objects. Since we do this aggregation in actual GPS space, we can cumulate likelihoods from different viewpoints, hence this treatment is theoretically invariant to camera movements, which often cause traditional methods, tracking features across frames, to fail (e.g. see Figure \ref{fig:heading_rotations}).

\begin{figure}
    \centering
    \includegraphics[width=1\textwidth]{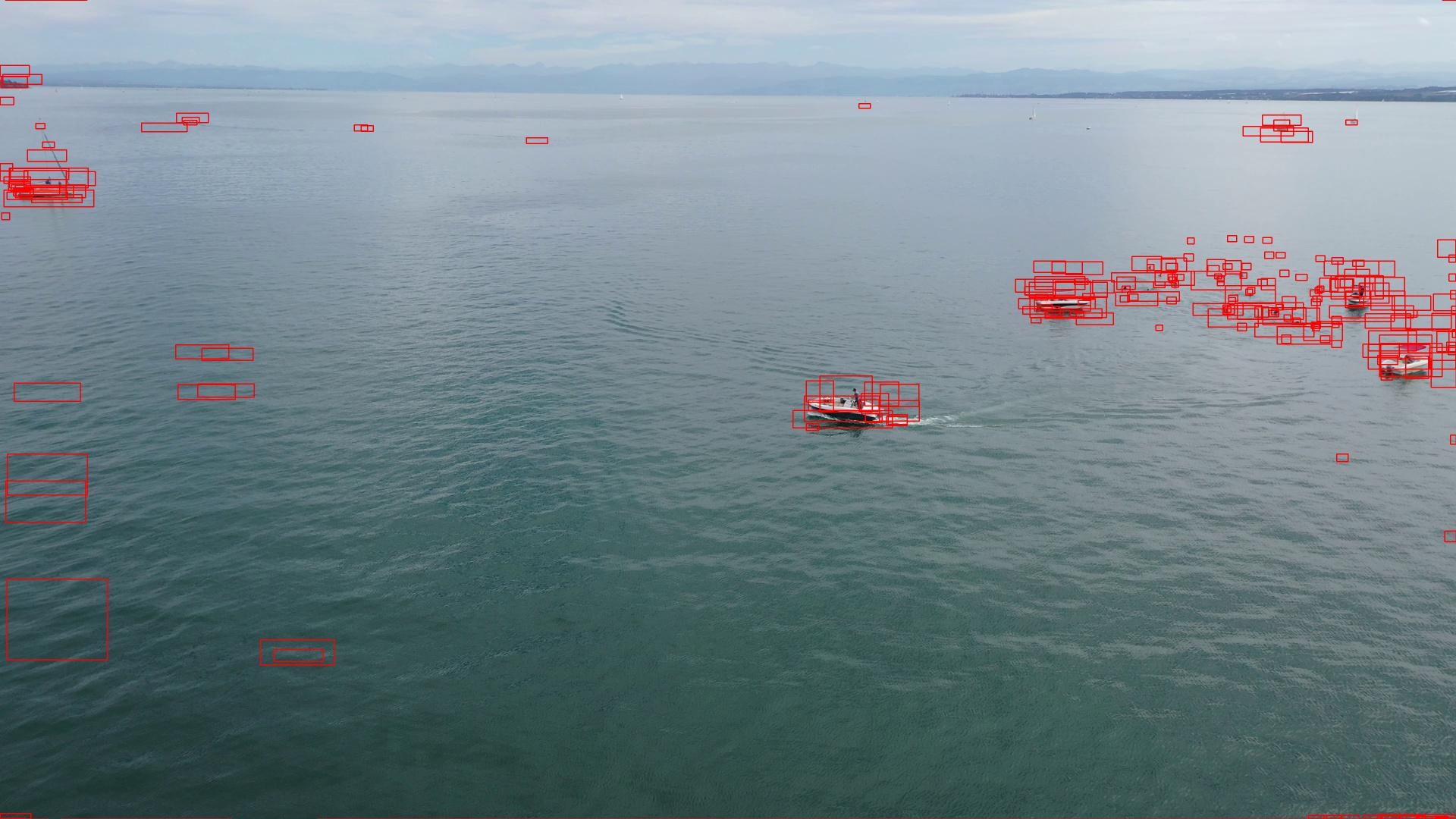}
    \caption{Before NMS, there are many low-confidence detections, but they cluster around actual objects. Ignoring the confidence, these detections would yield a recall value of one. For visualization, only one-colored boxes.}
    \label{fig:before_nms_example}
\end{figure}

To provide evidence for the initial claim about the high recall before NMS, we train a Faster R-CNN ResNet-18 and a YOLOv7 (configs from \cite{kiefer20231st}) on SeaDronesSee v2 \cite{kiefer20231st}. For evaluation, we remove the NMS stage and obtain recall values of 97.4 \% and 98.8\%, resp. Compare that to the recall after NMS stage: 86.4\% resp. 87.6\%. See also Figure \ref{fig:before_nms_example} for a sample image showing pre-NMS detections.

This leads us to hypothesize that a large part of missing detections can be obtained (and also false detections can be suppressed) if the confidence scores are modified such that low-confidence detections in \emph{GPS areas}, where there have been detections for several consecutive time steps, are boosted, and, simultaneously, (semi-)confident false positive detections are suppressed if they only appear in certain frames (randomly).

\textbf{Memory Map Construction:}

Therefore, we propose to map all detections before NMS to GPS space as follows. For every box detection $b_i=[x^i_1,x^i_2,y^i_1,y^i_2]$, $i=1, \dots, m$, we take its bounding box center $[(x^i_2+x^i_1)/2,(y^i_2+y^i_1)/2]$ and compute its GPS coordinates $g^{b_i}=[la^{b_i},lo^{b_i}]$ via the formulas in Section \ref{sec:mathsection}. We compute the closest grid window to $g^{b_i}$ via 
\begin{equation}
    w^{b_i} =  \argmin_{m\in M_t} ||  g^{b_i} - m ||.
\end{equation}

Ideally, this is the actual GPS position of that prediction. However, to account for imprecision in the sensor data and deviations from taking the center of the bounding box (which may not be the GPS center of the object), we assign weight to the neighbouring space as well, albeit less, since $w^{b_i}$ is our best single-point prediction\footnote{Naively applying this method on all boxes before filtering of incorrectly sized boxes leads to too large or small bounding boxes becoming boosted.}. Hence, we construct a memory map likelihood over $M_t$ as follows: We add a truncated Gaussian density with radius $r$ at $w^{b_i}$, i.e.,
\begin{equation}
    \Tilde{p}_t(x) \leftarrow \Tilde{p}_t(x)+s e^{-x^2} \text{ for } x\in {[w^{b_i}-r,w^{b_i}+r]\subset M_t},
\end{equation}
where we slightly abuse the notation by mixing scalars and vectors (computation to be understood component-wise). Note that we hid the other parameters (constants and variance) in the scaling factor $s$, which we will \emph{additionally} make dependent on the confidences $c_i$ belonging to $b_i$. 


After every time step, we truncate $\Tilde{p}$, such that the memory map values are $\leq 1$ at all times:

\begin{equation}
p_t(x) = \min (\Tilde{p}_t(x), 1).
\end{equation}
Furthermore, we introduce a forgetting factor $\phi$, by which the memory map $p_t$ is rescaled after each update step so that the memory map will not be overloaded over time.:
\begin{equation}
p_t(x) \leftarrow \phi p_t(x).
\end{equation}
Note that we explicitely did not model $p_t$ as a probability distribution over $M_t$ as this would introduce a competition of weights among the objects leading to deteriorated results when the number of objects and their prediction certainty vary (similar to a discrete Bayes filter \cite{fox2003bayesian}, although here, we do not have a control update).

\textbf{Adjusting the Detector Confidences According to $p_t$:}

At time $t$, each bounding box $b_i$ has a confidence $c_i\in[0,1]$, which we will update based on $p_{t-1}$ via
\begin{equation}
    \hat{c}_i=c_i + p_{t-1}(w^{b_i}).
\end{equation}
Note that we do not update the map based on the updated confidence $\hat{c}_i$ but on the original $c_i$. Lastly, note that we perform this procedure \emph{class-wise}, so we keep a map for each class separately. After the procedure, we proceed with standard NMS.

\subsection{Extension to Object Tracking and Reidentification}

Similarly as for VOD, a \emph{temporal memory map} helps in tracking scenarios. We apply the memory map to object tracking via a simple extension.

For frame $t$, we use the GPS location of each tracked object in frame $t-1$. We take the boosted predictions of frame $t$ closest to the old GPS location with a confidence above a threshold as the new location (in image and GPS space) of the object to be tracked. We will see that this simple formulation of a tracker helps to track objects over time in presence of camera movements, such as in Figure \ref{fig:heading_rotations}. Even if a fast camera movement yielding blurry frames results in the object detector missing the object temporally - if an object will be redetected in subsequent frames - the object can be assigned the correct ID again.  

While there is benefit in using memory maps for \emph{short-term} tracking tasks, they are especially beneficial in \emph{long-term} tracking, where ground truth objects may leave the frame entirely. The task of reidentification in classical scenarios is solved by feature-based comparison of objects across time. In UAV-based domains, this often fails for small objects as features look similar across different objects. However, the treatment in GPS space allows us to remember where an object was, allowing us to reidentify it in subsequent frames over a pre-defined time horizon.


\subsection{Extension to Video Anomaly Detection}

The goal of video anomaly detection is to output regions of the frames that are considered anomalous, either spatially (from appearance) or temporally (unseen movements). Since anomalies are not well-defined, methods achieve only limited success in completely detecting all anomalies, and the aerial nature leads to highly ambiguous scenes leading to many false positive detections as well \cite{kiefer2023fast}. Hence, this task is far from solved.

However, we may assume anomalies to be temporally stable, i.e. they exist over a considerable period of time. This makes it ideal to aggregate information at certain GPS positions over a sequence of frames. Since it is hard to explicitly define regions that are anomalous, most methods output an anomaly heatmap, indicating areas of likely anomalies. Therefore, we propose to map the entire image plane to GPS space and aggregate anomalous regions over time in a geometrically reasonable way. We note that this requires an efficient vectorized implementation to still achieve real-time inference. Section \ref{sec:video_anomaly_results} discusses this shortly.

For the correct mapping of pixels to GPS coordinates, we need to make sure that the projection is well-defined. Therefore, we apply the previously introduced horizon cutter, which first computes the horizon and only maps the pixels below it to GPS space. We ignore pixels too far in the horizon as this results in too much noise.

We replace the 2D difference heat map from \cite{kiefer2023fast} with our memory map, which will be added to the previous memory map. After each step, we also apply a forgetting factor $\phi$. Afterwards, we apply the same post-processing steps as indicated there.


\section{results and analysis}

\begin{figure}
    \centering
    \includegraphics[width=1\textwidth]{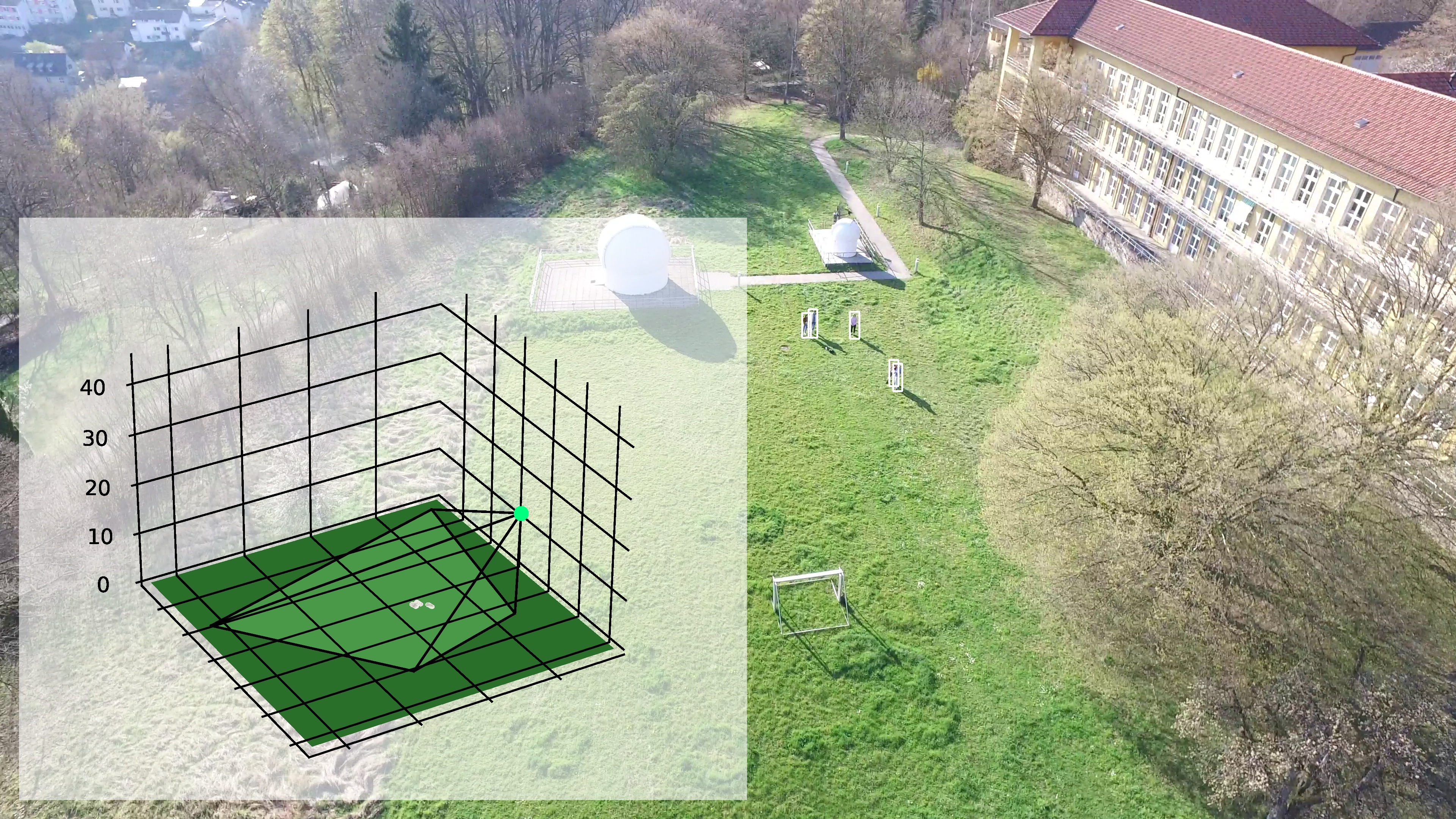}
    \caption{Example frame from our captured benchmark POG-V taken at $31$m altitude and a gimbal pitch angle of $41^\circ$.}
    \label{fig:example_pog}
\end{figure}


Since there is only one publicly available dataset for VOD and Multi-Object Tracking that employs precise metadata (SeaDronesSee-MOT \cite{varga2021seadronessee}), we collect and manually annotate another dataset, People-On-Grass-Video (POG-V), depicting people walking on grass and similar terrain (see example in Figure \ref{fig:example_pog}). It comes with the same type of metadata annotation as SeaDronesSee-MOT. 
In particular, we used a Zenmuse X5 camera mounted on a DJI Matrice 100 for collection. We collected 10,633 frames in 30 fps videos with 3840x2160 resolution. We annotated 48,802 instances of people using DarkLabel\footnote{\url{https://github.com/darkpgmr/DarkLabel}}. We made sure to have a great variance w.r.t. to altitude ($h=10m$-$100m$) and gimbal pitch angle ($\beta=17^\circ$-$90^\circ$).

Note that we focus on the detection of people in both benchmarks only, since this is the hardest class to detect \cite{kiefer20231st}. For that, we fuse the classes \emph{swimmers}, \emph{swimmers with life jacket} and \emph{life jacket} into a single people class in SeaDronesSee-V (SeaDronesSee-MOT without instance IDs).

For short-term tracking, we also employ SeaDronesSee-MOT. It currently only supports short-term tracking, i.e. objects that left the scene and reappear do not have the same id. Therefore, we relabel a video clip of SeaDronesSee-MOT, such that it also supports long-term tracking, which we need to assess the performance of long-term tracking (reidentification).

For Video Anomaly Detection, we fall back to the OpenWater data set of the Maritime Anomaly Detection Benchmark \cite{kiefer2021leveraging} since it is the only one featuring metadata.

Finally, we perform additional experiments to evaluate an extension to multiple UAVs in a collaborative scenario.


\subsection{Video Object Detection}

\begin{table}	
	\begin{center}		
		\begin{tabular}{lrrrrr}
                \toprule
                & \multicolumn{2}{c}{SeaDronesSee-V} & \multicolumn{2}{c}{POG-V}&\\
			Model name &  AP$_{50}$ & AR &  AP$_{50}$ & AR & FPS  \\
			\midrule
                DFF \cite{zhu2017deep} & 53.7 & 34.2 & 30.3 & 18.8 & 3.4  \\
                FGFA \cite{zhu2017flow} & 53.9 & 34.1 & 34.7 & 24.8 & 1.6\\
                T.R.A. \cite{gong2021temporal} & 60.6 & 42.9 & 29.9 & 19.3 & 2.4 \\
                YOLOv7Tiny & 68.2 & 41.1& 81.3 & 36.2 & \bf 25.1\\
                \bf +Mem. Map & \bf 72.8 & \bf 44.0 & \bf 84.4 & \bf 38.8 & \bf 25.1 \\
                \bottomrule
		\end{tabular}
	\end{center}
	\caption{Video Object Detection accuracy for Swimmers and People on SeaDronesSee-Video and POG-Video, respectively. The last column denotes running times (wall-clock time in frames per second) benchmarked on an Nvidia Xavier AGX.}
	\label{table:accuracy}
 \vspace{-3mm}
\end{table}

We apply the proposed memory map on the output of a YOLOv7-Tiny \cite{wang2022yolov7} trained and tested on SeaDronesSee-V and POG-V, respectively. We choose a grid with $0.5$m resolution and a size of $300$m. We truncate the Gaussian updates with a radius of $6$m. 

As baselines, we take the well-known Video Object Detectors Deep Feature Flow (DFF) \cite{zhu2017deep}, Flow-guided Feature Aggregation (FGFA) \cite{zhu2017flow} and the recent Temporal RoI Align (T.R.A.) \cite{gong2021temporal} as implemented in \cite{mmdetection} with their default configuration.

Table \ref{table:accuracy} shows that incorporating our memory maps leads to increased AP$_{50}$ and AR (average recall with averaged IoU levels 0.5:0.05:0.95 and at most 100 detections) values by successfully boosting the confidence scores. In particular, we achieve a +4.6 and +3.1  AP$_{50}$ increase over the single-image YOLOv7-Tiny object detector on SeaDronesSee-V and POG-V, respectively. Interestingly, popular video object detectors lack behind in performance because they are not targeted to aerial VOD. Moreover, the increased AR values shows that our method detects more objects, which is the main challenge in UAV-based detection.

\begin{figure}
    \centering
    \includegraphics[width=1\textwidth,trim={0 21 0 0},clip]{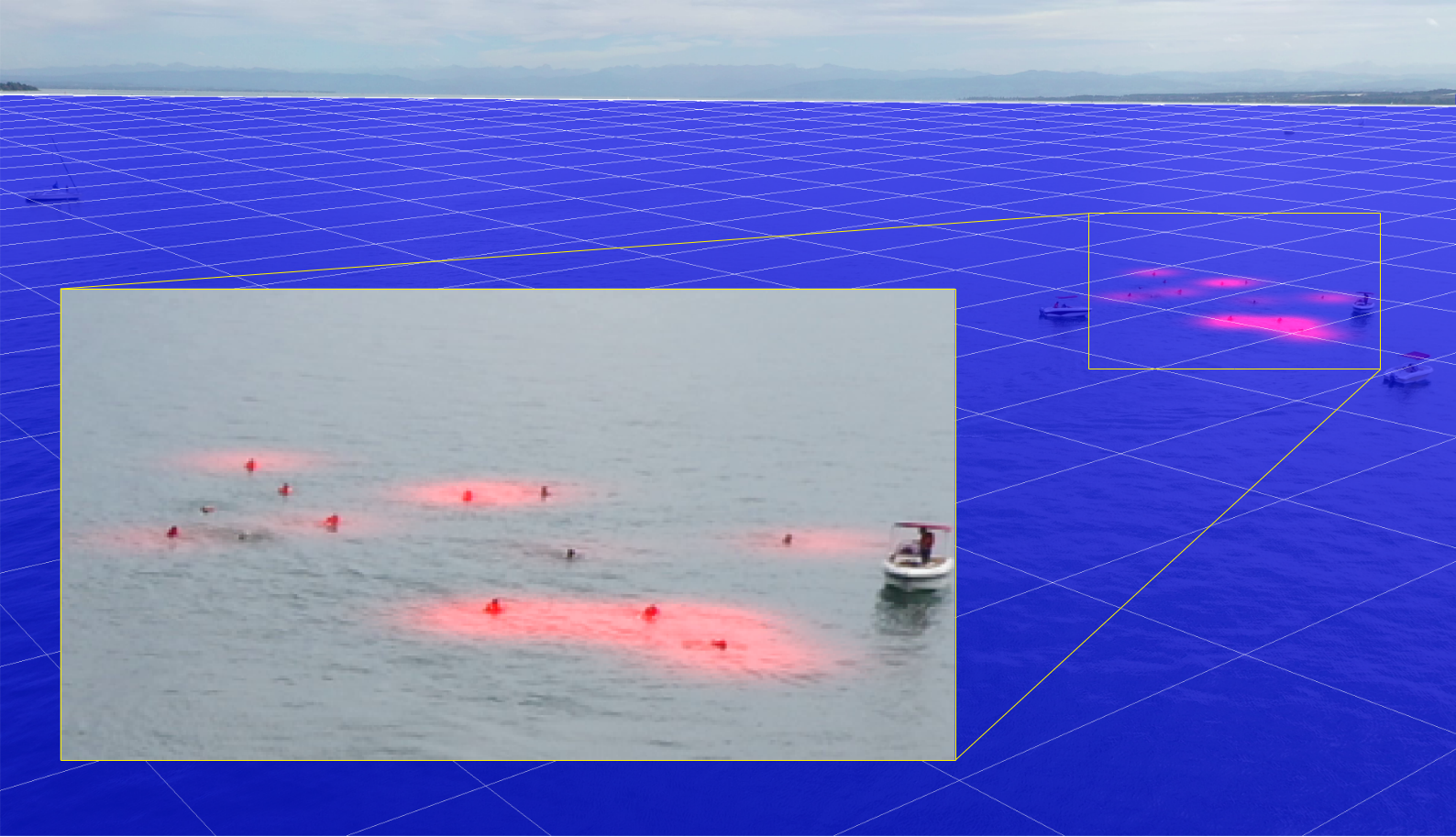}
    \caption{Heatmap illustration of memory map on a SeaDronesSee-V test frame. We project actual GPS latitude and longitude lines to image space. We also project the memory map from GPS space to image space and visualize it as a heatmap. The right image part shows that the memory map successfully puts weight around actual swimmers' locations. Compare also to the zoomed-in crop and Figure \ref{fig:intro_img} and \ref{fig:detected_objects_after_boosting}.}
    \label{fig:memory_map_inner_working}
\end{figure}

\begin{figure}
    \centering
    \includegraphics[width=1\textwidth,trim={0 20 0 0},clip]{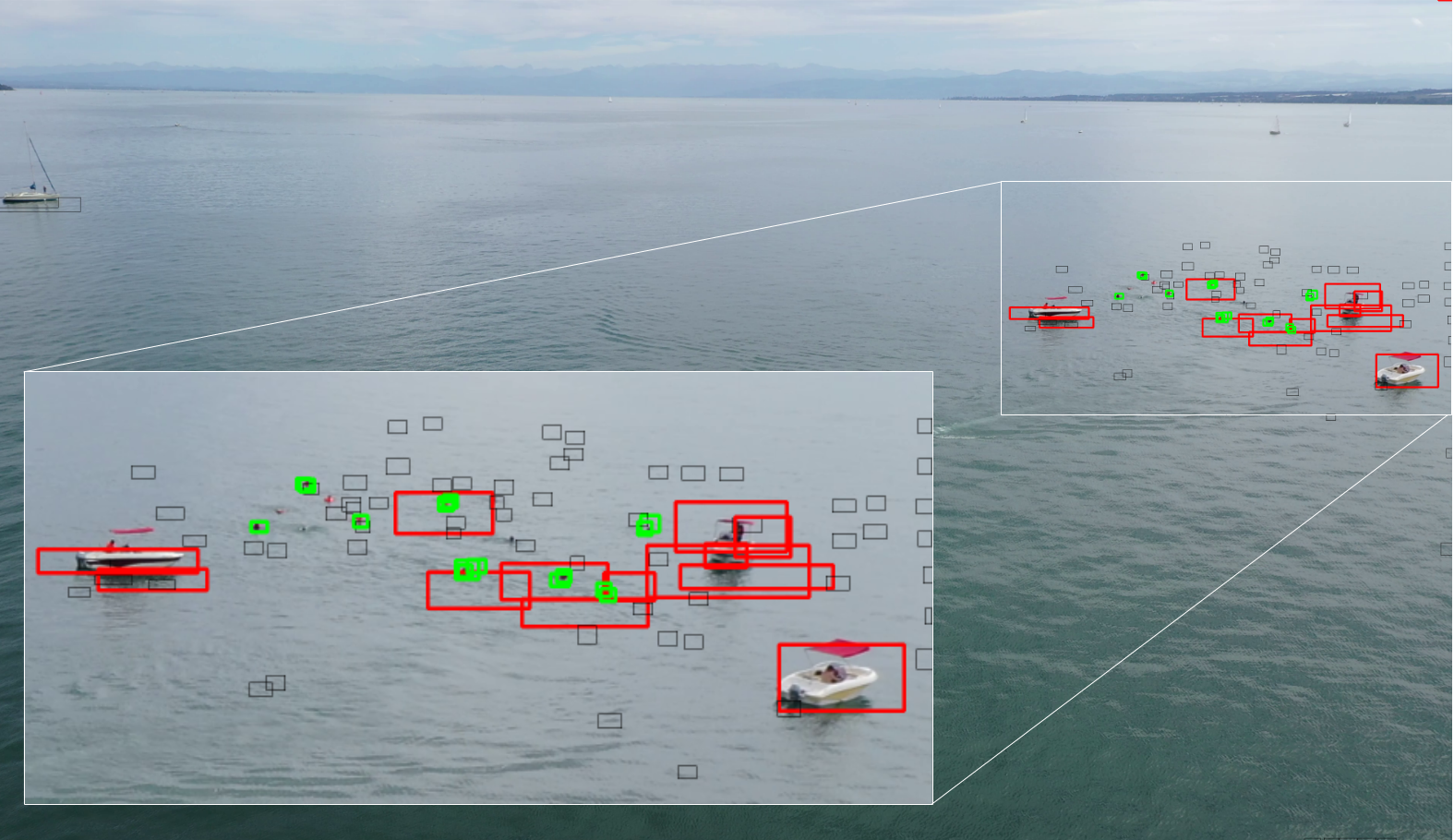}
    \caption{Before NMS, many swimmers are predicted (black boxes), several of which are automatically removed due to incorrect sizes (red boxes). Only the confidences of the green boxes are boosted. Applying a confidence threshold of 0.5 yields that 8 swimmers are detected as opposed to none for the baseline (experiment with YOLOv7-Tiny).}
    \label{fig:detected_objects_after_boosting}    
\end{figure}

Figure \ref{fig:memory_map_inner_working} visualizes the temporal memory map, projected to the image space. It shows the heatmap values of likely object locations, which were aggregated from the previous frames. The red areas cluster around actual objects. This results in pre-NMS boxes becoming boosted, as Figure \ref{fig:detected_objects_after_boosting} shows. While the baseline YOLOv7-Tiny originally did not detect a single swimmer, now we detect eight.

Furthermore, the standard video object detectors are not suitable for deployment on embedded devices. Our method only adds a negligible time to a standard YOLOv7-Tiny, which can run in real-time on an Nvidia Xavier AGX. 

\subsection{Object Tracking}

Applying our method on SeaDronesSee-MOT yields an increase in HOTA and MOTA and a decrease of ID switches and fragmentations over the DeepSORT short-term tracker (see Table \ref{table:accuracy_tracking}).

To test the reidentification capability, we run our method on top of a DeepSORT with reidentification module on the relabeled video as described earlier. We can successfully reassign the same id to all the objects in the video whereas the baseline fails to do so (4 people are reassigned new IDs.) 

\subsection{Video Anomaly Detection}
\label{sec:video_anomaly_results}

Table \ref{table:anomaly_detection} shows that adding the module to the anomaly detector from \cite{kiefer2023fast} yields an increase of +3 AR to 82.8 (average recall over multiple levels of broadcasting rate; measured differently than average recall from object detection, please refer to  \cite{kiefer2023fast}). This performance increase is also reflected in the recall at broadcasting rate of 5$\%$, i.e. there is a +3.8 R$^{p=5\%}$ improvement over the Autoencoder.

\begin{figure}
    \centering
    \includegraphics[width=1\textwidth,trim={0 0 0 0},clip]{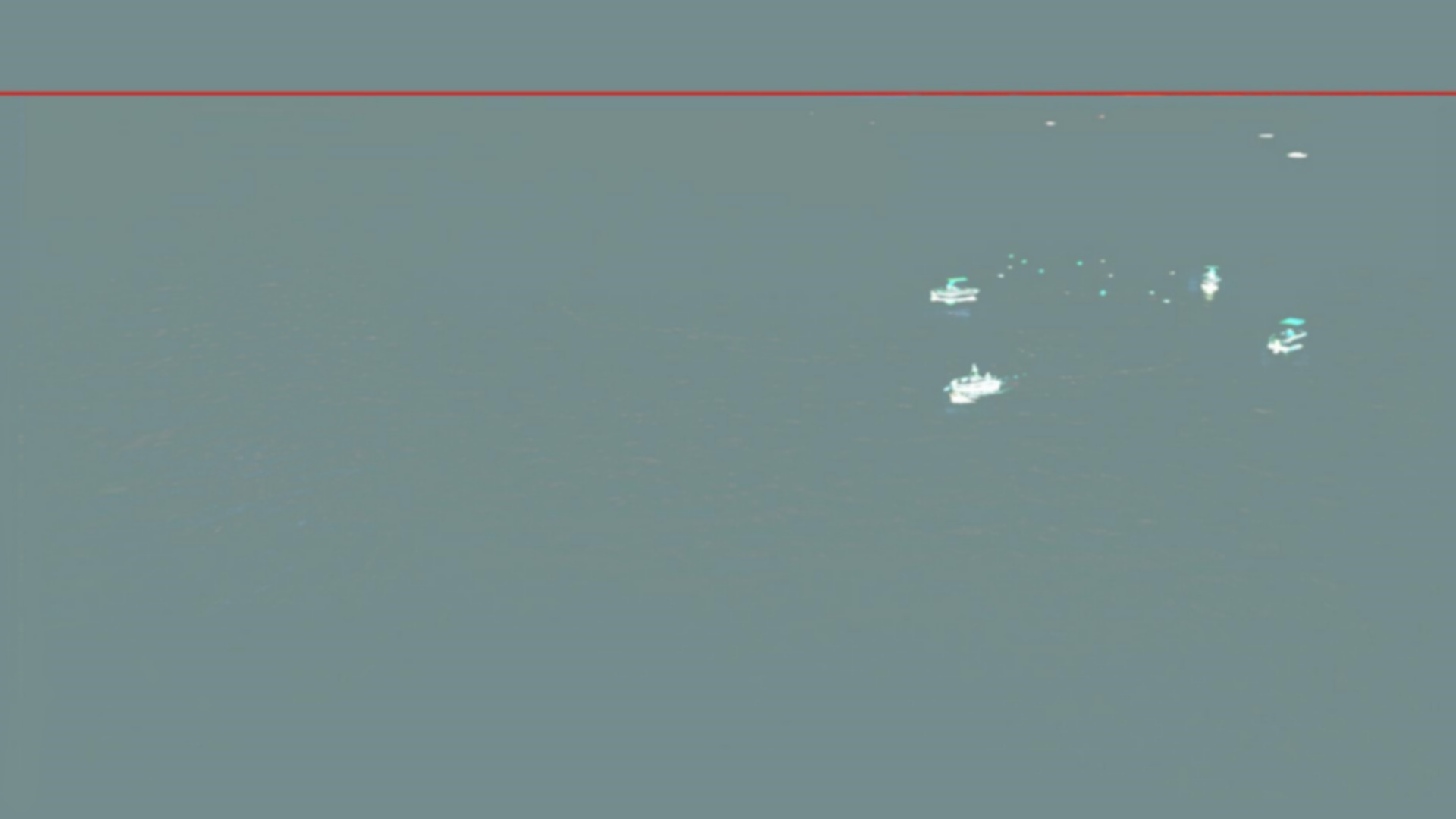}
    \includegraphics[width=1\textwidth,trim={0 0 0 0},clip]{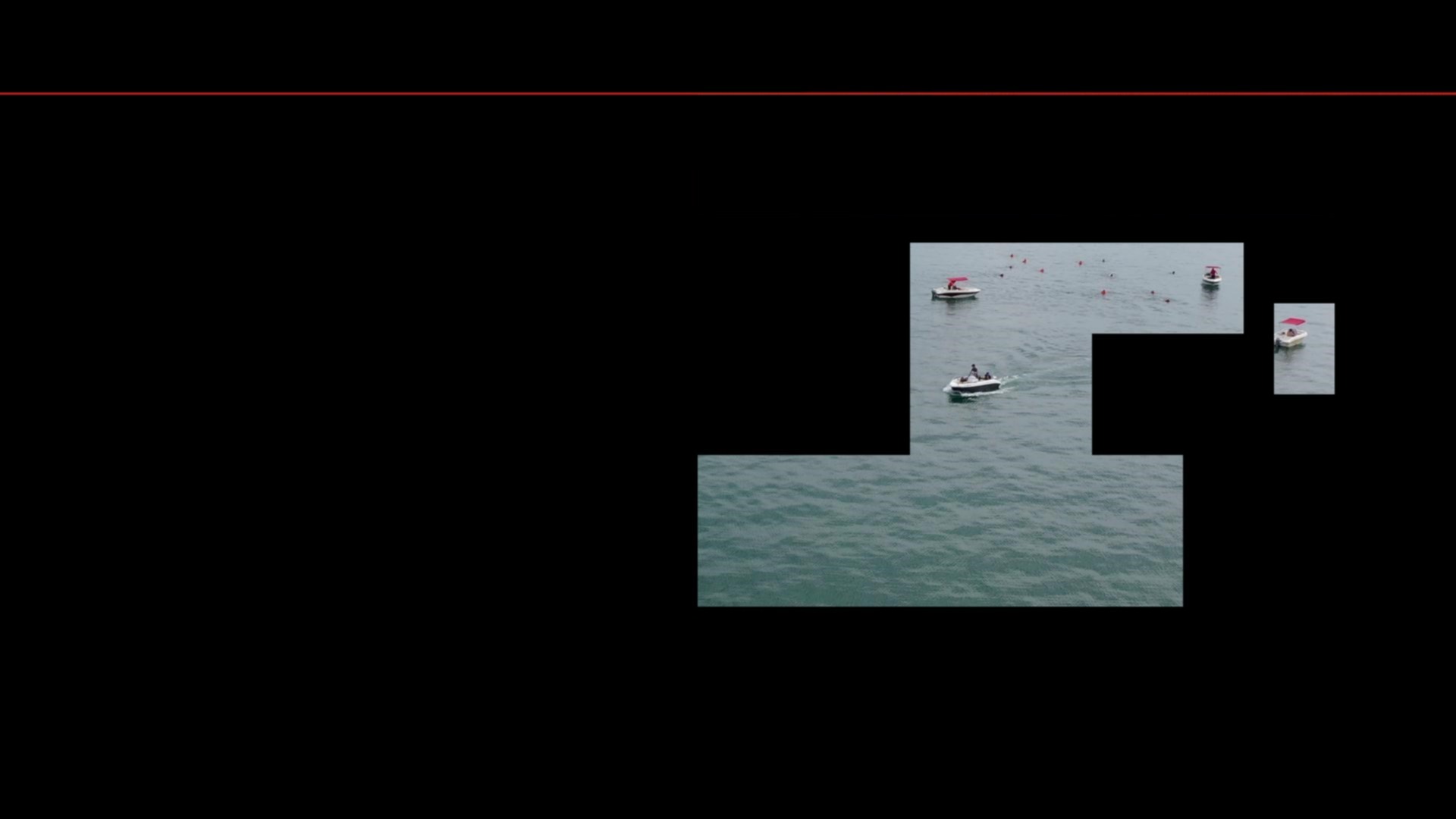}
    \caption{Difference anomaly map aggregated in 3D space over several frames (top) and resulting regions of interested (bottom).}
    \label{fig:autoencoder}
\end{figure}

Figure \ref{fig:autoencoder} shows the resulting difference anomaly map aggregated over the last frames and the resulting regions of interest that are returned. 

The speed does not increase considerably over the Autoencoder, still achieving real-time inference benchmarked on an Nvidia Xavier AGX as Table \ref{table:anomaly_detection} shows.

\begin{table}	
	\begin{center}		
		\begin{tabular}{lrrrr}
                \toprule
			Model name &  HOTA$\uparrow$ & MOTA$\uparrow$ &  IDs$\downarrow$ & Frag$\downarrow$  \\
			\midrule
                ByteTracker \cite{kiefer20231st} & 65.0 & 76.9 & 68 & 841   \\
                DeepSORT \cite{kiefer20231st} & 66.6 & 80.0 & 44 & 805 \\
                \bf +Memory Map & \bf 67.2 & \bf 80.8 & \bf 35 & \bf 721\\
                \bottomrule
		\end{tabular}
	\end{center}
	\caption{Multi-Object Tracking accuracy on SeaDronesSee-MOT. We used the output of DeepSORT and built on top a mem. map to become more robust towards id switches and fragmentations.}
	\label{table:accuracy_tracking}
    \vspace{-3mm}
\end{table}

\begin{table}	
	\begin{center}		
		\begin{tabular}{lrrr}
                \toprule
			Model name &  AR$\uparrow$ & R$^{p=5\%}\uparrow$ & FPS  \\
			\midrule
                Gaussian Mixture Model \cite{kiefer20231st} & 45.9 & 2.6 & 17 \\
                Mean Filter \cite{kiefer20231st} & 73.9 & 54.3 & 50 \\
                Frame Differencing \cite{kiefer20231st} & 76.1  & 54.8 & 62   \\
                Autoencoder \cite{kiefer20231st} & 79.8 & 71.0 & 27  \\
                \bf +Memory Map & \bf 82.8 & \bf 74.8 & 27\\
                \bottomrule
		\end{tabular}
	\end{center}
	\caption{Video Anomaly Detection accuracy on SeaDronesSee-MOT. We built our memory maps on top of the Autoencoder \cite{kiefer2023fast}.}
	\label{table:anomaly_detection}
\vspace{-3mm}
\end{table}

\begin{figure*}
    \centering
    \includegraphics[width=0.36\textwidth,cframe=red]{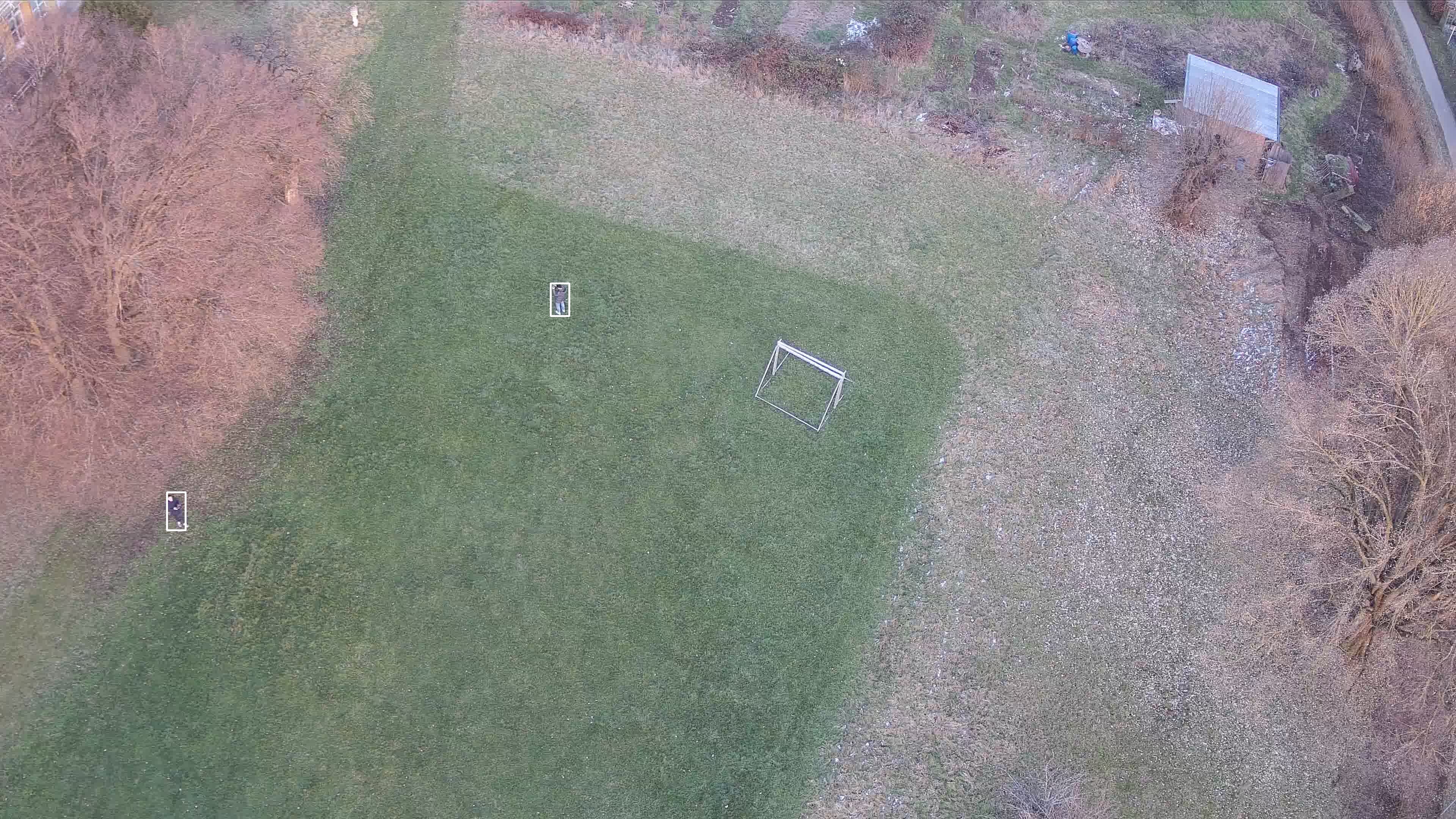}
    \includegraphics[width=0.25\textwidth]{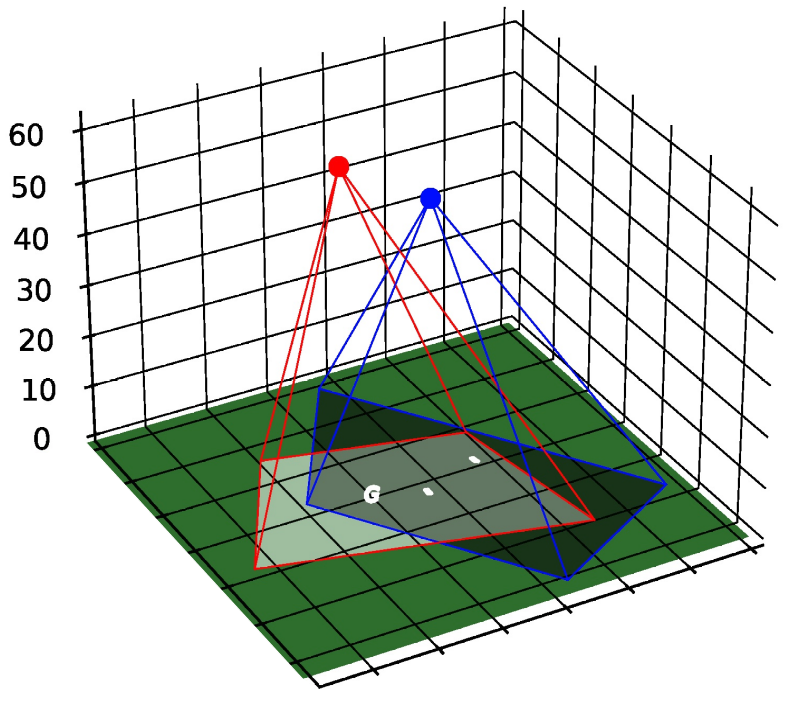}
    \includegraphics[width=0.36\textwidth,cframe=blue]{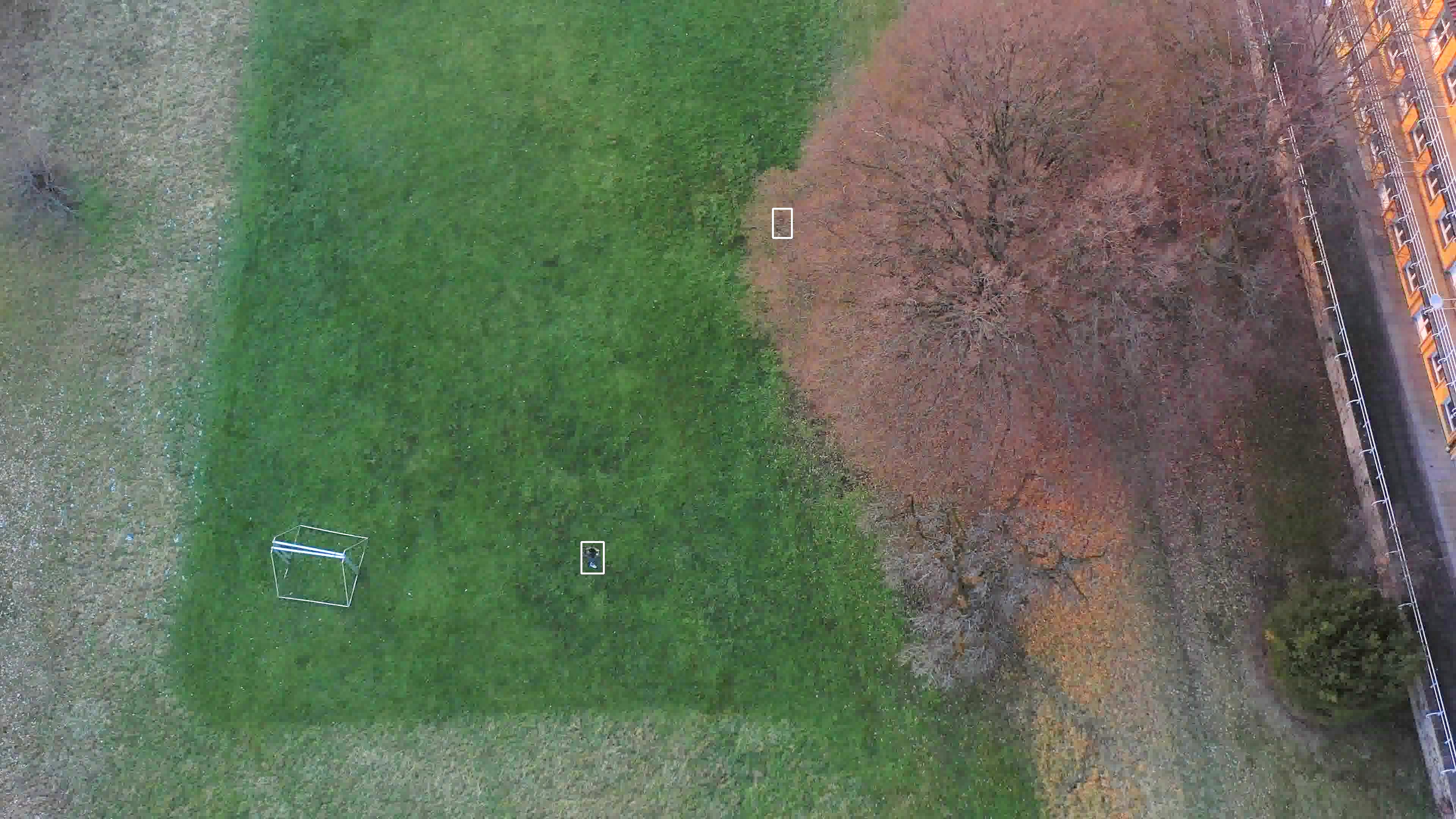}

    \caption{Two UAVs inspect the same scenery from different locations at the same time. The red camera (left) has a good view on both objects and the goal (G) while the blue camera (right) is underexposed (brightened for visualization) and one person is partially occluded by a tree. We aggregate both memory maps together to obtain more robust detections. Without considering the underlying geometry, this is not possible with standard video object detectors. }
    \label{fig:two_UAVs}
\end{figure*}

\subsection{Cooperative Detection via Multiple UAVs}

The GPS memory map allows for a joint representation of objects' locations. In this section, we demonstrate that it can be leveraged in a collaborative setting with multiple UAVs. This allows for cross-UAV knowledge transfer and leads to more robust predictions. Furthermore, it allows for detections of (partially) occluded objects. For example, Figure \ref{fig:two_UAVs} shows the same scenery captured from different locations at the same time with two people are walking on the grass.

Having a joint location likelihood representation allows the UAVs to share information. We apply the same memory map from before on both UAVs' object detectors' output, but this time, we average their memory maps in the overlapping region. For that, we take an EfficientDet-D$0$ trained on POG \cite{kiefer2021leveraging} and test it on the following data. We capture four minutes of footage in a similar environment as POG from the viewpoint of two UAVs, a DJI Matrice 210 and a DJI Mavic 2 Pro, denoted \emph{2AVs}. We varied variables, such as altitude, pitch and heading viewing angle and GPS location. We annotated the people visible in both video streams and compare the performance of the single-frame object detector with the memory map from before. We obtain an AP$_{50}$ of 61.3 for the single-frame object detector compared to an AP$_{50}$ of {\bf 68.9} for the memory map. 

To shot the utility of a joint memory map for tracking, we compare a fast single-object tracker, PrDiMP18 \cite{danelljan2020probabilistic}, to a simple tracker based on the memory map. For that, we apply an EfficientDet-D$0$ on a consecutive subset of 2AVs that contains partial occlusions of a person in one video stream (blue camera in Fig. \ref{fig:two_UAVs}).  While the baseline tracker cannot handle the occlusion and fails to track the person behind the tree entirely (4.3 AP$_{50}$) for the blue camera, our method leverages the joint memory map that transfers knowledge from the red UAV to the blue (86.3 AP$_{50}$).

The joint memory map also allows for reidentification in long-term tracking tasks \emph{while the object is moving}. To test this, we take a standard DeepSORT trained on POG with the default Reid model within mmtracking \cite{mmdetection} as baseline. We take a subset of 2AVs where one camera leaves the scenery entirely, while the other is tracking the object throughout. When reappearing, it immediately uses the track id information from the other camera, while DeepSORT failed to reidentify the object, which we believe to be attributed to the object size.

\section{Conclusion and Discussion}

While not using meta-data remains the standard in computer vision on UAVs, using metadata to boost the performance shows promising results without inducing a large computational overhead. We showed that we can improve standard methods by considering the underlying 3D geometry. Reasoning about detections, tracklets and anomalies in an interpretable and robust way allows for more trustworthy methods.


Although in our experiments on POD-Video we did not encounter any problems with terrain elevation change, future work remains to show how strict the assumption of even-level terrain is on other benchmarks. For that, it is inevitable to collect larger and more diverse benchmarks.

In future works, it will be interesting to see how a symbiosis of metadata and other computer vision tasks looks like. Lastly, it seems relevant to dive into the topic of uncertainty quantification methods to account for the errors in metadata values.

\bibliographystyle{IEEEtran}
\bibliography{IEEEabrv,IEEEexample}

\end{document}